\documentclass[12pt,a4paper]{article}
\usepackage[top=1in,bottom=1in,left=1in,right=1in]{geometry}

\usepackage{amsmath,amssymb,nccmath}

\usepackage{changepage}

\usepackage[utf8x]{inputenc}

\usepackage{textcomp,marvosym}

\usepackage{cite}

\usepackage{nameref,hyperref}

\usepackage{microtype}
\DisableLigatures[f]{encoding = *, family = * }

\usepackage[table]{xcolor}

\usepackage{array}

\raggedright
\usepackage[aboveskip=1pt,labelfont=bf,labelsep=period,justification=raggedright,singlelinecheck=off,font=small]{caption}

\usepackage{lastpage,fancyhdr,graphicx}
\usepackage{epstopdf}
\fancyheadoffset[L]{2.25in}
\fancyfootoffset[L]{2.25in}

\usepackage[ruled,vlined,noend]{algorithm2e}
\usepackage{algorithmic}
\usepackage{mathtools}
\usepackage{siunitx}
\usepackage{setspace}
\usepackage{ragged2e}

\begin{document}
\vspace*{0.2in}
\onehalfspacing

\begin{flushleft}
{\Large
\textbf\newline{Forest Fire Clustering for Single-cell Sequencing with Iterative Label Propagation and Parallelized Monte Carlo Simulation} 
}
\newline
\\
Zhanlin Chen\textsuperscript{1},
Jeremy Goldwasser\textsuperscript{1},
Philip Tuckman\textsuperscript{2},
Jason Liu\textsuperscript{3},
Jing Zhang\textsuperscript{4,*},
Mark Gerstein\textsuperscript{1,3,5,*}
\\
\bigskip
\textbf{1} Department of Statistics and Data Science, Yale University, New Haven, CT 06520, USA
\\
\textbf{2} Department of Earth, Atmosphere, and Planetary Sciences, Massachusetts Institute of Technology, Cambridge, MA 02139, USA
\\
\textbf{3} Department of Molecular Biophysics and Biochemistry, Yale University, New Haven, CT 06520, USA
\\
\textbf{4} Department of Computer Science, University of California, Irvine, CA 92617, USA
\\
\textbf{5} Department of Computer Science, Yale University, New Haven, CT 06520, USA
\\
\bigskip

%
%





* jing.zhang@uci.edu, pi@gersteinlab.org

\end{flushleft}

\justifying


\section*{Abstract}

In the era of single-cell sequencing, there is a growing need to extract insights from data with clustering methods. Here, we introduce Forest Fire Clustering, an efficient and interpretable method for cell-type discovery from single-cell data. Forest Fire Clustering makes minimal prior assumptions and, different from current approaches, calculates a non-parametric posterior probability that each cell is assigned a cell-type label. These posterior distributions allow for the evaluation of a label confidence for each cell and enable the computation of ``label entropies," highlighting transitions along developmental trajectories. Furthermore, we show that Forest Fire Clustering can make robust, inductive inferences in an online-learning context and can readily scale to millions of cells. Finally, we demonstrate that our method outperforms state-of-the-art clustering approaches on diverse benchmarks of simulated and experimental data. Overall, Forest Fire Clustering is a useful tool for rare cell type discovery in large-scale single-cell analysis. 

\section*{Introduction}

Clustering analysis is an important statistical method with many application scenarios. In single-cell sequencing, clustering analysis groups individual cells into subtypes, such as categorizing subtypes of cancer cells for targeted therapy \cite{tang2009mrna, RN13}. However, single-cell sequencing measurements are sparse and uncertain, and solely categorizing cells into different groups is not enough to develop a statistical view of the cell identity \cite{kharchenko2021triumphs}. In addition to clustering, we should also quantify the confidence of cell-type labels assigned to each cell. Indeed, one of the biggest challenges in single-cell sequencing analysis is validating a computational analysis method \cite{kiselev2019challenges}. 

Most current clustering methods can be roughly grouped into five general categories: centroid-based, distribution-based, connectivity-based, density-based, and graph-based methods \cite{saxena2017review}. Centroid-based clustering methods, such as K-means clustering, rely on optimizing a central vector to find data clusters \cite{RN20}. In distribution-based clustering methods, clusters are defined as samples from certain distributions (e.g., Gaussian distributions in Gaussian mixture models) \cite{reynolds2009gaussian}. Connectivity-based clustering methods, also known as agglomerative or hierarchical clustering, iteratively merge data points into the same group based on linkage to form a hierarchical structure \cite{RN19}. Similar to linkages in connectivity-based methods, density-based clustering methods, such as DBSCAN and OPTICS, produce clusters based on reachability \cite{RN23, RN22}. Graph-based methods, such as Louvain and Leiden, maximize the modularity to find communities in the data graph (e.g., the K-nearest neighbor [KNN] graph)  \cite{de2011generalized, RN24}. These methods can discover the number of communities or clusters in the data indirectly specified by the resolution hyperparameter. 

Overall, a suitable clustering algorithm for single-cell data should have these three important features:

\begin{enumerate}

  \item A single-cell clustering algorithm should make relatively weak assumptions about the data (e.g., the shape of the data) for rare cell type discovery. Each clustering method makes certain assumptions about the data distribution, in part reflected by their hyperparameters. The number and strength of the assumptions influence the clustering results. A number of well-known clustering methods (such as K-means and Gaussian mixture models) make relatively strong prior assumptions about the data, which can bias and mislead investigators to overlook rare but important cell types. For example, K-means is biased towards identifying equal-sized clusters, which may result in rare cell types being hidden among a larger group \cite{kiselev2019challenges}. Indeed, previous works have suggested that these rare cell types could challenge many of our assumptions about data distribution and cellular composition \cite{buettner2015computational, lake2018integrative}. Comparatively, the resolution parameter in Louvain is a weaker assumption than the K in K-means. Therefore, a clustering method should have a small number of hyperparameters while making minimal assumptions about the data. 

  \item A single-cell clustering method should have the ability to internally validate its clustering results. Currently, only hierarchical clustering can perform internal validation by testing whether a particular cluster exists in the data using bootstrap resampling techniques \cite{van2003new}. Potentially, a more interpretable and useful statistic is the probability of having the label at each point. In single-cell analysis, this probability could be understood as a confidence measure for the cell type of each cell. 
  
  \item A single-cell clustering algorithm should be computationally efficient. As the number of cells sequenced continues to grow, single-cell RNA-sequencing (scRNA-seq) datasets can have more than a million cells, and clustering once on such a large dataset can take days \cite{abdelaal2019comparison}. Therefore, it is important to design a scalable clustering algorithm for large single-cell datasets.
\end{enumerate}

To meet these specifications, we developed Forest Fire Clustering inspired by self-organized criticality in forest fire dynamics. By modeling label propagation similar to the spread of forest fires, we can cluster data given only a ``fire temperature" hyperparameter, similar to the resolution hyperparameter in Louvain. Through simulating label propagations from different starting points, we can compute point-wise posterior exclusion probabilities (PEP), similar to p-values, to quantify the probability that a data point takes on the other cluster labels. We can also compute the point-wise label information entropy to measure the general uncertainty of the labels at each data point. Lastly, Forest Fire Clustering can be used for online clustering, which is a useful feature for analyzing high-throughput data. As experimental data accumulate over time or as data stream over a network, the computational cost of re-clustering continues to rise as the number of cells sequenced increases. Due to the inductive nature of the algorithm, Forest Fire Clustering can make online inferences on a small number of newly arrived data points without re-clustering.

Here, we provide an overview of Forest Fire Clustering and show how it performs against other clustering methods. However, it is difficult to evaluate the performance of our algorithm because ground truth labels are hard to obtain for unsupervised tasks. Therefore, we applied two general strategies to benchmark the accuracy, efficiency, and robustness of Forest Fire Clustering. First, we simulated synthetic and scRNA-seq data from known distributions with ground truth labels, and we compared clustering labels to ground truth labels using metrics such as the adjusted Rand index (ARI) and the purity score. Second, we benchmarked Forest Fire Clustering against other clustering methods on different single-cell sequencing technologies, tissues and data scales with labels generated by marker genes, cell surface antibodies, jointly profiled scATAC-seq (single-cell Assay for Transposase-Accessible Chromatin followed by sequencing), and subject matter experts (Supplementary Figure 1, 2). With these strategies, we demonstrate the utility of Forest Fire Clustering in discovering rare cell types, highlighting novel transition cells in developmental pseudo-time, and making accurate online inferences.

\section*{Results}

The main idea of Forest Fire Clustering is to envision data points as trees in a forest and cluster labels as fires that repeatedly propagate through the forest. Modeling forest fire dynamics generates interpretable clusters by iteratively propagating labels through the data manifold. To minimize prior assumptions, the algorithm has only one effective hyperparameter that indirectly governs the number of clusters discovered. Additionally, a point-wise posterior label distribution for internal validation is constructed by simulating label propagations from random starting points via Monte Carlo.

\subsection*{Method Overview}

There are three main steps in the Forest Fire Clustering algorithm: 

\begin{enumerate}
	\item Preprocessing: The data matrix $\pmb W$ with rows as cells and columns as genomic features is used to compute the cell-to-cell pairwise $\mathbb{L}^2$ distance matrix $\pmb M$. Then, $\pmb M$ is transformed into an affinity matrix $\pmb A$ using the kernel method (Fig~\ref{figure1}a, Step 1-2). The affinity matrix serves as the adjacency matrix of the data graph, with data points as vertices and similarities as edges. With only local affinities, the data graph connects data points to reflect global relationships and represents the data geometry in a low-dimensional manifold. The data graph can be approximated with a KNN graph, and an adaptive kernel can be used to accurately capture both the local structure and the long-range interactions between data points. \useshortskip
	
	\item Label Propagation: On the data graph, a random unlabeled vertex $r$ is selected as a seed to take on a new label (Fig~\ref{figure1}a, Step 3). Each unlabeled vertex $i$ has a label acceptance threshold $T_i$, which is defined as the inverse of its degree. In dense regions of the graph, where the vertices have higher degrees, the inverse relationship results in lower label acceptance thresholds. Thus, highly connected vertices can accept and propagate labels more easily, and vertices in dense regions tend to be in the same cluster. The labeled seed $r$ radiates label influence $H_{r, i}$ (or heat) to all other unlabeled vertices $i$ computed by multiplying the user-defined ``fire temperature" parameter $c$ with the affinity $A_{r, i}$. Hence, the farther the unlabeled vertices are from the seed, the less label influence they experience. If the average label influence $\bar H_{labeled, i}$ from all labeled vertices on the unlabeled vertex $i$ exceeds the acceptance threshold $T_i$, then vertex $i$ takes on the label of the seed (Fig~\ref{figure1}a, Step 4-5). Even if $\bar H_{labeled, i}$ is not high enough to cross the threshold $T_i$, it is possible to cross the threshold as more vertices are labeled later on (Fig~\ref{figure1}a, Step 6-7). Therefore, Forest Fire Clustering checks the remaining unlabeled vertices each time new vertices are labeled until the average label influence cannot exceed the threshold of any unlabeled vertices (Fig~\ref{figure1}a, Step 8). \useshortskip
	
	\item Iterative Label Propagation: New label propagations are iteratively performed until all vertices have been labeled, and each round of label propagation defines a cluster (Fig~\ref{figure1}a, Step 9-10). Between rounds of label propagation, labeled vertices from the previous rounds are considered to be exhausted and can no longer be replaced by new labels. By doing so, nonlinear clusters in the data manifold can be found by propagating labels in the data geometry. Sparse regions in the data manifold form natural cluster boundaries, which can be outlined by iterative label propagation. 

\end{enumerate}

\begin{figure}[!h]
\centerline{\includegraphics[width=\columnwidth]{formatted_figures/figure1.png}}
\caption{{\bf Illustration of Forest Fire Clustering and Monte Carlo Validation:} a) Illustration of Forest Fire Clustering. In the data preprocessing stage, a KNN data graph is created by transforming the pairwise distances into affinities using adaptive kernels (Steps 1-2). Then, a vertex is randomly selected as the seed to take on a label (Step 3). The label is allowed to propagate to other vertices by determining if the average label influence crosses the threshold at each vertex (Steps 3-8). Lastly, the process is iteratively repeated until all vertices have been given a label (Steps 9-10). b) Pseudo-code for the Forest Fire Clustering algorithm. c) Posterior label distributions are computed for each data point. Labels are repeatedly propagated from random seed vertices with the same set of parameters. For each point, the labels over all Monte Carlo trials are collected to construct an empirical posterior label distribution, which can be used to compute the posterior exclusion probability and the label entropy.}
\label{figure1}
\end{figure}

	We can evaluate the confidence of Forest Fire Clustering labels using Monte Carlo simulations (Fig~\ref{figure1}c). After performing Forest Fire Clustering, we store the initial clustering results $S_{1...n}$ for all $n$ vertices. Then, a posterior label distribution for each data point can be constructed using label permutations. In each permutation, a random seed vertex $i$ is selected, and the seed vertex takes on the same label $S_i$ as it did in the initial clustering results. Labels on all other vertices are cleared. The seed label $S_i$ is allowed to propagate until the average label influence can no longer cross the threshold of the remaining unlabeled vertices. Vertices that accepted the seed label $S_i$ add one occurrence of that seed label in their records. We repeat with another randomly chosen seed vertex $j$ with a seed label $S_j$, which could come from a different initial cluster than the previous seed ($S_j \neq S_i$). After many Monte Carlo trials, the accepted seed labels are tabulated at each data point to construct a discrete label frequency distribution, which we call the posterior label distribution. 
	
	Similar to the posterior distribution in Bayesian inference, the posterior label distribution is an updated probability distribution over the labels of a data point after accounting for new information from the Monte Carlo simulations. The posterior exclusion probability, which quantifies label confidence using the posterior label distribution, is defined as the posterior probability that a data point accepts the other labels rather than the initial label. Similar to p-values, the posterior exclusion probability is a simulated, retroactive probability that the initial label was not accepted across all Monte Carlo trials. Having a high posterior exclusion probability means that the initial label is less likely to be the true label for the data point, whereas a low value means that the initial label is more robust. The information entropy of the posterior label distribution can be used to measure general uncertainty of the labels at each data point. Additionally, simulating many Monte Carlo trials can be efficiently parallelized because each trial is independent of one another.
	
	Clustering labels can be extended to a small number of new data points in an online-learning context (Supplementary Figure 3). As a new data point arrives, we can iteratively check whether label influences from existing clusters can cross the threshold of the new vertex. If multiple label influences from different clusters exceed the threshold, then the new data point takes on the label with the highest average influence. If none of the average label influences can cross the threshold, then the data point becomes the seed vertex for a new cluster. Thus, in contrast to previous online clustering methods, Forest Fire Clustering can discover new clusters in newly arrived data, which reduces the computational cost for re-clustering and scales with the growth of high-throughput technologies. 

\subsection*{Evaluating Forest Fire Clustering on Synthetic Data}

\begin{figure}[!h]
\centerline{\includegraphics[width=\columnwidth]{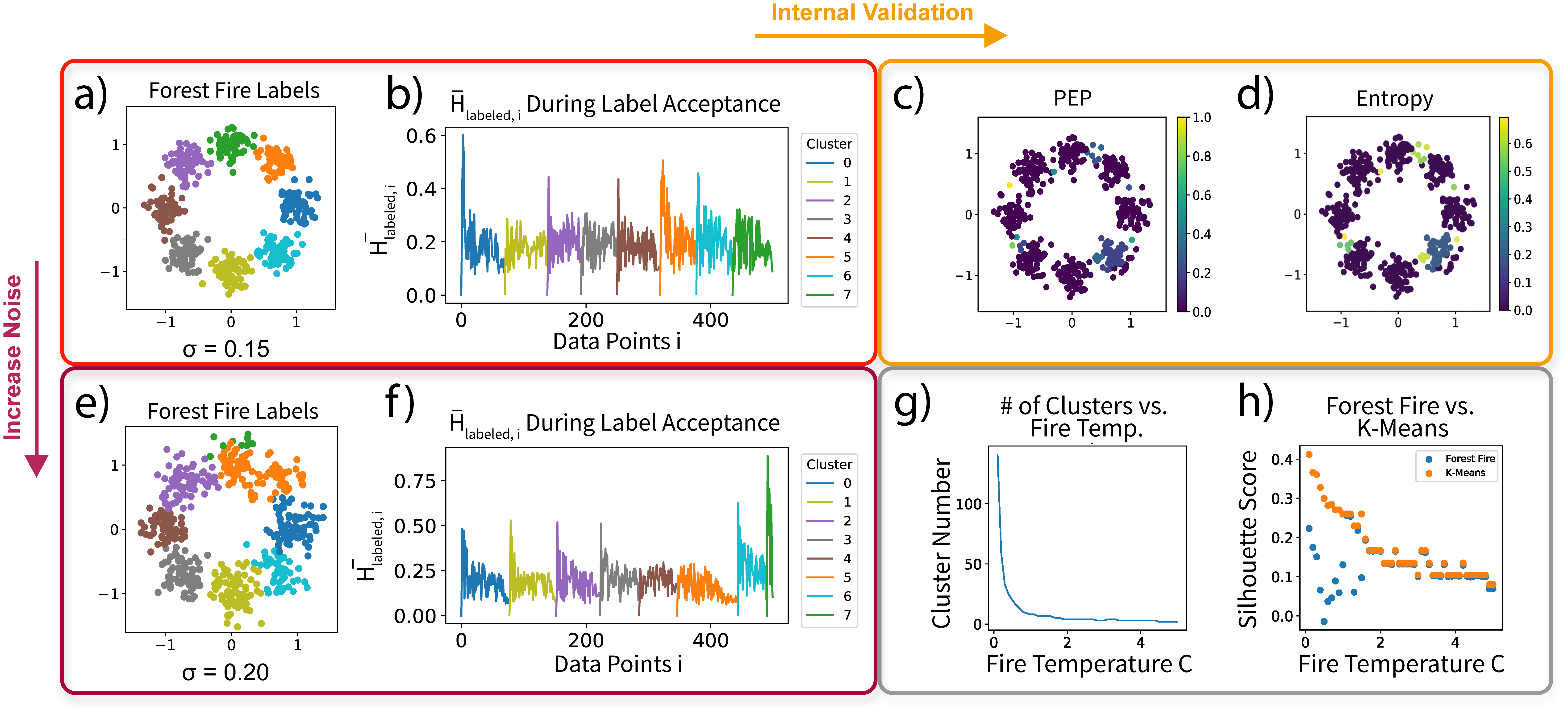}}
\caption{{\bf Visualizing the Forest Fire Clustering Process with Gaussian Mixture Models:}
a) Forest Fire Clustering labels on Gaussian mixtures ($\sigma=0.15$) centered on the unit circle. b) Average heat $\bar H_{labeled, i}$ during label acceptance plot when $\sigma=0.15$. c) Posterior exclusion probability (PEP) of the Forest Fire labels on Gaussian mixtures from part (a). d) Entropy of the Forest Fire labels on Gaussian mixtures from part (a). e) Forest Fire Clustering labels on Gaussian mixtures ($\sigma=0.20$) centered on the unit circle. f) Average heat $\bar H_{labeled, i}$ during label acceptance plot when $\sigma=0.20$. g) The number of clusters found in the data decreases as the fire temperature increases. h) Silhouette score of Forest Fire Clustering vs. K-means clustering on Gaussian mixtures from part (a). }
\label{fig2}
\end{figure}

We first examined the performance of Forest Fire Clustering on simulated Gaussian mixtures (Fig~\ref{fig2}a, e). We constructed two Gaussian mixture models ($\sigma=0.15$ and $\sigma=0.2$) with eight distributions located evenly around the unit circle ($n=500$). Forest Fire Clustering identified a cluster corresponding to each Gaussian in the mixture. As fire temperature $c$ indirectly determines the cluster sizes, Forest Fire Clustering can infer the number of clusters in the data by separating different Gaussians with the sparse regions in between.

Interestingly, the number of clusters can also be inferred from the number of peaks in the average label influence plot (Fig~\ref{fig2}b; right). This plot records the average label influence $\bar H_{labeled, i}$ of each vertex $i$ when it accepts a label, which traces the dynamics of the clustering process and provides a new perspective for interpreting the clustering results. As the cluster size increases, the unlabeled vertices become farther away from the cluster center and experience less average heat. With a low Gaussian variance, every spike corresponds to a meaningful cluster that was discovered. With a higher Gaussian variance, eight clusters were discovered, although two clusters were separated differently from the ground truth (shown in green and orange). Accordingly, the average label influence over time plot indicated different spikes for those clusters (Fig~\ref{fig2}f). Therefore, these plots enhance interpretability and could serve as a heuristic in determining the quality of the clustering results. 

To internally validate the previous clustering results, we constructed posterior label distributions using Monte Carlo simulations, and we computed the posterior exclusion probability and label entropy of each data point in the Gaussian mixtures (Fig~\ref{fig2}c, d). The validation shows higher label entropies over a particular cluster and between a few clusters. Further, data points on the boundaries of clusters have higher posterior exclusion probabilities, indicating that Forest Fire Clustering is less confident about the labels on those data points. In single-cell analysis, the posterior exclusion probabilities can be used to control the quality of the cell types for each cell. 

In addition to conducting internal validation, we evaluated the quality of the clusters discovered by Forest Fire Clustering. By varying our only effective hyperparameter $c$, we investigated whether the algorithm could consistently find high-quality clusters. We verified our hypothesis that as the fire temperature $c$ increases, the average cluster size also increases (Fig~\ref{fig2}g). Therefore, the fire temperature $c$ is an intuitive parameter used for generating clusters of different sizes. In addition, we discovered the number of clusters $K$ using Forest Fire Clustering and compared with K-means using the same $K$ (Fig~\ref{fig2}h). As the fire temperature $c$ increases, the silhouette score of Forest Fire Clustering converges to that of K-means. Since the silhouette score of K-means is locally optimal given the number of clusters K, this suggests that Forest Fire Clustering can generate approximately optimal clusters with an appropriate fire temperature $c$. In the Methods section, we further explain the ability for Forest Fire Clustering to generate high-quality clusters without an objective function and demonstrate the asymptotic convergence of Forest Fire clusters to true cluster distributions.

We performed simulated experiments to examine whether Forest Fire Clustering can infer the emergence of new clusters in out-of-sample data. With only four training clusters ($\sigma=0.1$), Forest Fire Clustering demonstrated strong performance as the number of unseen clusters in the out-of-sample data increased, with average ARI and purity score both greater than 0.90 throughout (Supplementary Figure 4). Additionally, the out-of-sample cluster discovery remains robust with increasing Gaussian noise. However, Forest Fire Clustering can only guarantee consistently high online-learning performance when there is more training data than test data, because large training datasets create well-defined manifolds that are less impacted by various test data arrival sequences. 

\subsection*{Comparing Forest Fire Clustering with Other Clustering Methods}

\begin{figure}[!h]
\centerline{\includegraphics[width=\columnwidth]{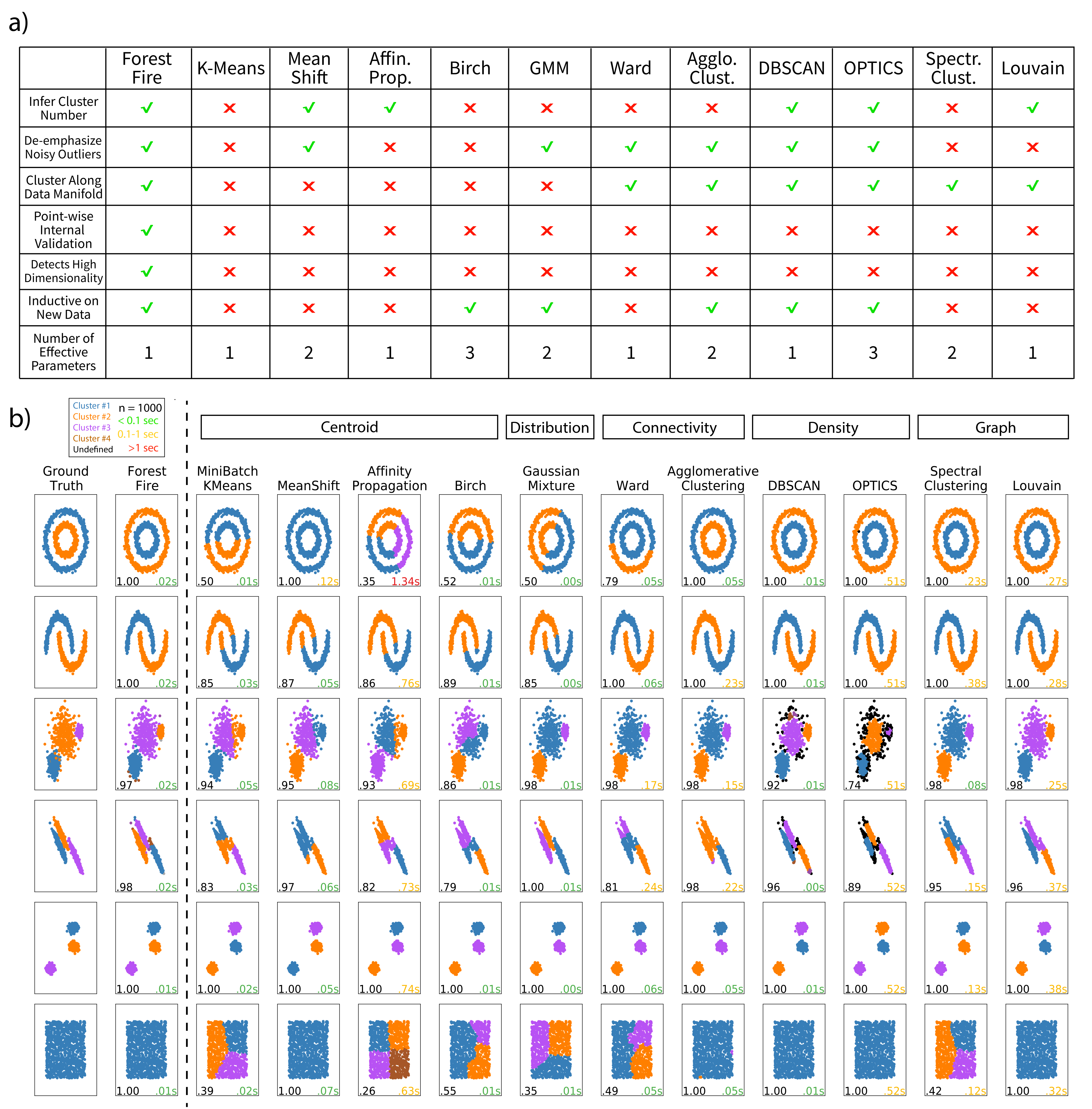}}
\caption{{\bf Comparing Forest Fire Clustering with Other Clustering Methods:}
a) Table comparing features of Forest Fire Clustering with many existing clustering methods. b) Benchmarking Forest Fire Clustering with other types of clustering methods on synthetic datasets (n = 1000). Lower left denotes the purity score, and lower right records the runtime (green $\leq 0.1$s; yellow $> 0.1$s and $\leq 1$s; red $> 1$s). }
\label{fig3}
\end{figure}

As mentioned previously, Forest Fire Clustering is designed to overcome weaknesses of many clustering methods on single-cell sequencing datasets (Fig~\ref{fig3}a). When compared with other clustering methods, Forest Fire Clustering can generate clusters with minimal prior assumptions and can compute a non-parametric point-wise posterior probability for internal validation. In addition to these unique advantages, we benchmarked Forest Fire Clustering with many existing clustering methods on a large number of synthetic datasets (Fig~\ref{fig3}b, Supplementary Figure 5, 6). 

For ring-shaped or half-moon-shaped data (Fig~\ref{fig3}b), distribution-based methods like Gaussian mixture modeling could not accurately categorize these datasets and achieved a relatively low purity score of 0.5 and 0.85, respectively. Both distribution-based and centroid-based methods are limited in identifying non-convex clusters. In contrast, the cluster boundaries from our method are more flexible because label propagation in the data manifold makes minimal assumptions about the shape of the data. Moreover, the results suggest that Forest Fire Clustering can better uncover the number and size of the clusters than connectivity-based methods, especially when the intra-cluster distance is larger than the inter-cluster distance. Similar to the linkage function in hierarchical clustering, Forest Fire Clustering propagates its labels to the closest data points when the fire temperature is above a certain threshold. In essence, both methods cluster by locally linking closest data points into the same group. 

Our method also maintains robust performance when the clusters overlap, whereas density-based methods like DBSCAN and OPTICS perform poorly in this scenario. In particular, given three overlapping clusters in the data, the $eps$ hyperparameter in DBSCAN and OPTICS must be sufficiently small to discover the three clusters, but a small $eps$ hyperparameter reduces the reachability and misclassifies some data points between clusters as singleton clusters (with singleton points in black). In contrast, Forest Fire Clustering can be seen as a continuous version of DBSCAN. The smooth non-linear Gaussian kernel allows for a more flexible representation of the data manifold and serves as a better alternative than the hard $eps$ cutoff. Compared to Spectral Clustering, Forest Fire Clustering can find the number of clusters in the data more accurately. Because label propagation only labels each data point once, Forest Fire Clustering also has a significantly shorter runtime compared to Louvain.

\subsection*{Evaluating Forest Fire Clustering on Simulated scRNA-seq Data}

SC3 and Seurat are considered popular and widely used single-cell clustering methods \cite{duo2018systematic, stuart2019comprehensive, kiselev2017sc3}. In particular, SC3 performs hierarchical clustering on a consensus matrix constructed by repeated K-means clustering on different partitions of the Laplacian eigenvectors, and Seurat maximizes the modularity in a KNN data graph using Leiden. We compared the performance of Forest Fire Clustering with SC3 and Seurat on simulated scRNA-seq data. Using the Splatter package in R \cite{zappia2017splatter}, we generated two types of simulated data for benchmarking: a mixture of discrete cell populations ($n=20$ datasets) and continuous stem cells along paths of differentiation ($n=20$ datasets).

\begin{figure}[!ht]
\centerline{\includegraphics[width=\columnwidth]{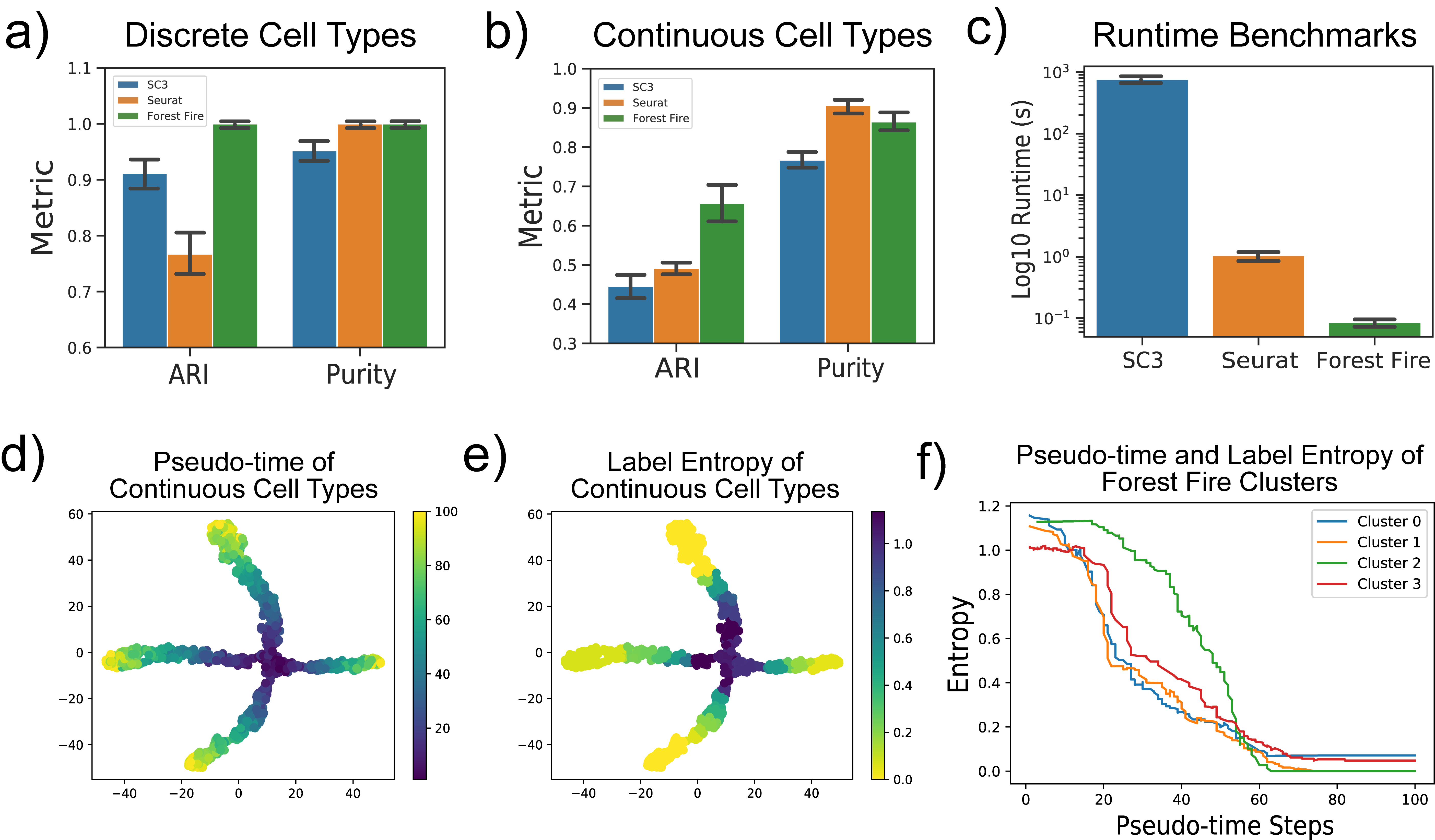}}
\caption{{\bf Simulated scRNA-seq Analysis:} 
a) Discrete cell types were simulated with Splatter ($n=20$ independent datasets) and used to benchmark SC3, Seurat (Louvain), and Forest Fire Clustering. With five discrete cell types, Forest Fire Clustering outperforms SC3 and Seurat in ARI and SC3 in purity score. b) Continuous cell types were simulated with Splatter ($n=20$ independent datasets) and used to benchmark SC3, Seurat (Louvain), and Forest Fire Clustering. With four cell developmental paths, Forest Fire Clustering outperforms SC3 and Seurat in ARI and SC3 in purity score. c) Forest Fire Clustering outperforms SC3 and Seurat in runtime ($n=20$ independent datasets). Data are presented as mean values +/- 2SD. d-e) For continuous cell paths, a stem cell type differentiates into four cell types. As pseudo-time step increases, each cell type becomes more specialized. Forest Fire entropy is able to mimic the pseudo-time step along the data manifold. f) Within each cluster, label entropy negatively correlates with pseudo-time. }
\label{fig4}
\end{figure}

For simulated discrete cell types (Fig~\ref{fig4}a), Forest Fire Clustering outperformed SC3 in terms of ARI and purity scores (Wilcoxon signed-rank  $p=9.5 \times 10^{-7}$). Forest Fire Clustering also outperformed Seurat in ARI (Wilcoxon signed-rank $p=7.7 \times 10^{-5}$) but showed no statistically significant difference in purity score. We observed a similar performance in ARI and purity score for continuous cell types (Fig~\ref{fig4}b). This indicates that both Forest Fire Clustering and Seurat can generate homogenous clusters, yet Forest Fire Clustering is better at correctly discovering the intrinsic number of cell types and subcommunities. Further, Forest Fire Clustering outperformed SC3 and Seurat in runtime with an average of $\SI{80}{\milli\second}$ compared to $\SI{12}{\minute}$ and $\SI{1}{\second}$, respectively (Fig~\ref{fig4}c). 

Moreover, the internal validation reveals that the label entropy of a cell negatively correlates with the pseudo-time steps along the paths of differentiation (Fig~\ref{fig4}e, f). As novel transition cells differentiate and become more specialized along the developmental trajectories, the identities of the cells also become more well defined (Fig~\ref{fig4}g). Point-wise label entropy through Forest Fire internal validation can capture the change in the uncertainty of the cell identity: transition cells tend to have high label entropy while differentiated cells tend to have low label entropy. Therefore, Forest Fire Clustering can highlight novel transition populations in developmental pseudo-time and lend deeper insights into single-cell analyses compared to many existing clustering methods.

\subsection*{Evaluating Forest Fire Clustering on PBMC Single-cell Data}

To demonstrate the advantage of Forest Fire Clustering on experimental data, we analyzed single-cell sequencing of around 10,000 peripheral blood mononuclear cells (PBMCs) from two datasets. In the first dataset, PBMCs were sequenced with CITE-seq, which measures single-cell gene expression with cell-surface proteins levels \cite{stoeckius2017simultaneous}. We used signature cell-surface proteins to categorize PBMCs into various cell types and clustered each cell based on gene expression. In the second dataset, multiomic sequencing jointly profiled the single-cell chromatin accessibility (scATAC-seq) and gene expression \cite{reyes2019simultaneous}. We labeled the cells using PBMC marker genes and clustered each cell based on the ATAC peaks. For benchmarking, the Louvain method is a popular single-cell clustering algorithm that maximizes modularity, and Leiden improves upon Louvain in runtime and finding well-connected communities. PARC is a fast combinatorial graph clustering method that incorporates pruning before applying Leiden \cite{stassen2020parc}. 

\begin{figure}[!ht]
\centerline{\includegraphics[width=\columnwidth]{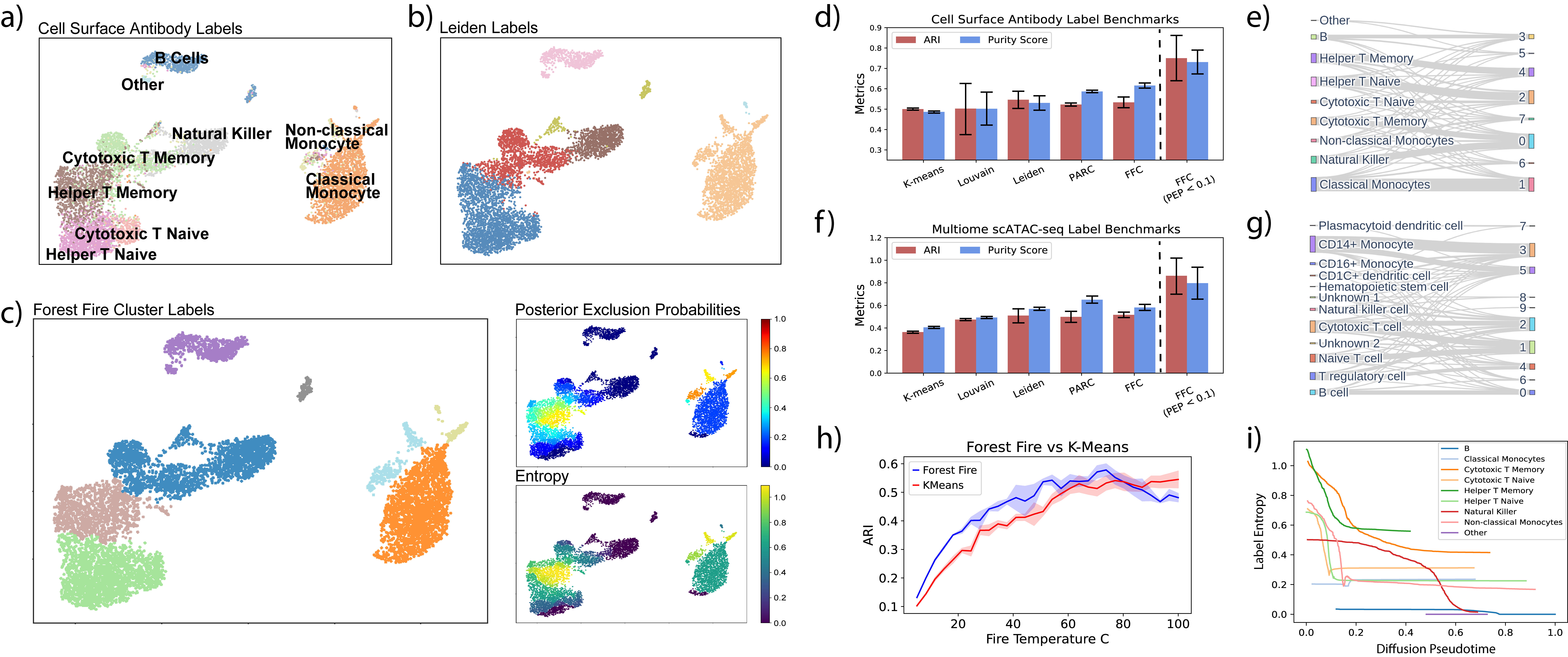}}
\caption{{\bf Benchmarking Forest Fire Clustering on PBMC data:} 
a) Ground truth cell-surface antibody labels. b) Leiden cluster labels. c) Forest Fire clusters with point-wise posterior exclusion probabilities and label entropies from internal validation. d) Clustering quality benchmarks using labels generated by cell-surface protein levels while clustering the gene expression ($n=20$ different seeds). e) Sankey diagram of surface protein labels and Forest Fire clusters on gene expression. f) Clustering quality benchmarks using labels generated by marker genes while clustering the ATAC peaks ($n=20$ different seeds). Data are presented as mean values +/- 2SD. g) Sankey diagram of marker-gene labels and Forest Fire clusters on ATAC peaks. h) When Forest Fire Clustering was used to discover the number of clusters $K$, the cluster quality with Forest Fire Clustering is often better than the cluster quality with K-means (using the same number of clusters $K$ and $n=20$ different seeds). Since Forest Fire Clustering outperforms K-means over a wide range of fire temperatures, the results also indicate that Forest Fire Clustering can discover high-quality clusters at different cluster resolutions. i) Point-wise label entropy versus diffusion pseudo-time. Within most surface-protein-labeled cell types, the label entropy decreases as diffusion pseudo-time increases. }
\label{fig5}
\end{figure}

The results suggest that our method can correctly categorize major PBMC cell types (Fig~\ref{fig5}). Clustering quality benchmarks indicate that Forest Fire Clustering can consistently produce clusters with similar ARI and purity scores compared with other state-of-the-art single-cell clustering methods (Fig~\ref{fig5}d, f). In addition, Forest Fire Clustering can discover high-quality clusters at different cluster resolutions and cell population sizes (Fig~\ref{fig5}g, Supplementary Figure 7). For online clustering, each cell type was ablated from the training data, and batch effects were corrected using Mutual Nearest Neighbors correction \cite{RN204}. Clustering quality metrics suggest robust performance in discovering new cell types in the testing data (Supplementary Figure 8, 9). 

We then evaluated the effect of internal validation on the Forest Fire Clusters. By focusing on cells with high-confidence labels (PEP $< 0.1$), we can improve the cluster ARI by more than $20\%$ compared to existing methods (Fig~\ref{fig5}b, d, Supplementary Figure 10). When analyzing discrete cell types, our internal validation can robustly discover prototypical cells, which could potentially benefit downstream analyses such as differential expression and cell-to-cell communication network modeling. Similar clustering benchmarks on human cortex scRNA-seq and mouse skin multiomic sequencing further support these conclusions (Supplementary Figure 11, 12).

Additionally, Forest Fire Clustering can analyze continuous cell types. Recapitulating results from our simulation studies, the label entropies decreased as diffusion pseudo-time increased within each cell type, and progenitor cells in diffusion pseudo-time analysis matched the cells highlighted by Forest Fire label entropy \cite{haghverdi2016diffusion} (Fig~\ref{fig5}e). However, as a top-down approach, diffusion pseudo-time analysis risks making incorrect assumptions about the signature genes expressed in transitional cells, which could lead to the wrong choice of root for diffusion. Instead, Forest Fire Clustering adopts a bottom-up approach by empirically uncovering cells with high differentiation potency along the data manifold. Compared with diffusion pseudo-time, our method has the unique advantage of assuming no prior knowledge about novel transition cells \textit{a priori} (e.g., signature gene expression) \cite{RN202}. By initiating label propagation from various cells in their developmental trajectories, Forest Fire can find novel transition cell types at the intersections of these label propagations where the label entropy is high. We further demonstrated this capability with scRNA-seq data on mouse hematopoietic stem cells (mHSCs) \cite{paul2015transcriptional} and human embryonic stem cells (hESCs) spanning physiological developmental days \cite{moon2019visualizing} (Supplementary Figure 13). Therefore, our method can naturally uncover branching points along developmental trajectories and reveal new biological insights from single-cell sequencing.

\subsection*{Evaluating Forest Fire Clustering on Large-scale Mouse Single-cell Data}

As the number of cells sequenced by single-cell technologies grows, the scalability of clustering algorithms becomes increasingly important. Here, we demonstrate the efficiency of Forest Fire Clustering compared with other state-of-the-art clustering algorithms on large heterogeneous datasets. Specifically, the Mouse Cell Atlas (MCA) contains gene expression data of around 400,000 cells from various mouse tissue profiled using Microwell-seq \cite{han2018mapping}. Compared to other large-scale sequencing datasets, the tissue of origin for cells in MCA can serve as ground truth labels for comparisons among clustering algorithms.

\begin{figure}[!ht]
\centerline{\includegraphics[width=\columnwidth]{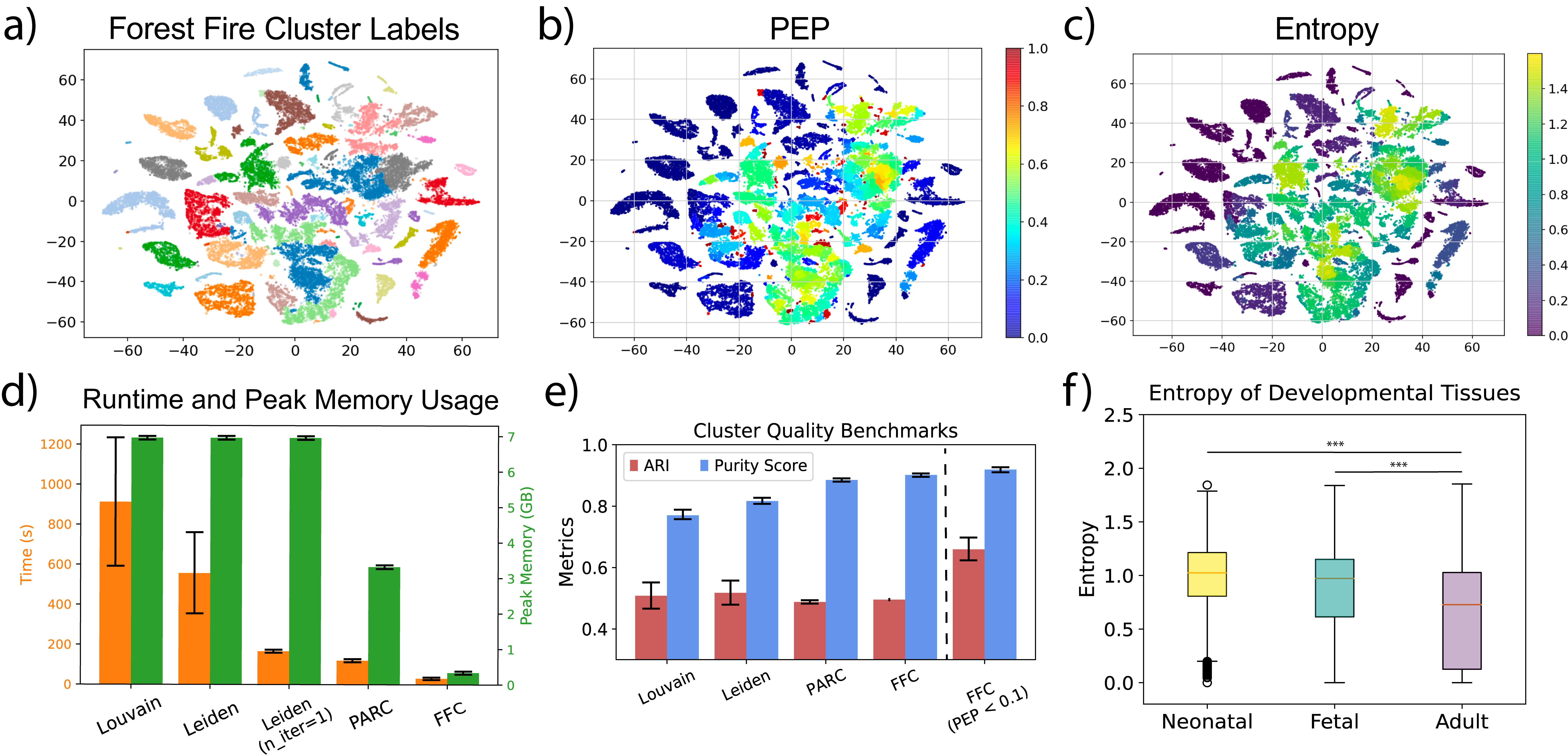}}
\caption{{\bf Benchmarking Forest Fire Clustering on MCA data:} 
a) Forest Fire cluster labels. b) Posterior exclusion probability for the Forest Fire cluster labels. c) Entropy for the Forest Fire cluster labels. d) Clustering runtime and peak memory usage benchmarks ($n=20$ different seeds). e) Clustering quality benchmarks ($n=20$ different seeds). Note that cells with high-confidence labels (PEP$<0.1$) from the internal validation significantly improved the clustering quality. Data are presented as mean values +/- 2SD. f) Label entropy at different developmental stages. The box plot visualizes data based on the minimum, first quartile or 25 percentile as lower box bound, median as center, third quartile or 75 percentile as higher box bound, and maximum. *** denotes two-sided Mann-Whitney U tests with $p < 1 \times 10^{-3}$.}
\label{fig6}
\end{figure}

The results indicate that Forest Fire Clustering is faster than state-of-the-art single-cell clustering algorithms while generating similar quality clusters (Fig~\ref{fig6}d, e). Additionally, Forest Fire Clustering uses less memory compared to existing methods. The runtime and memory usage benchmarks suggest that Forest Fire Clustering can efficiently scale to larger datasets. As expected, benchmarking on 1.3 million mouse brain cells from 10x Genomics supported these findings (Supplementary Figure 14).

Next, we investigated the effect of internal validation on Forest Fire clusters (Fig~\ref{fig6}e). For a heterogeneous dataset like MCA, focusing on cells with high label confidence (PEP $<0.1$) improved the ARI from 0.38 to 0.72. Additionally, neonatal and fetal mouse tissues exhibit higher label entropies, supporting our previous conclusion that label entropies mimic developmental time (Fig~\ref{fig6}f). Further, cells with high label entropies differentially express developmental and cell differentiation marker genes (Supplementary Figure 15, 16). However, computationally intensive Monte Carlo simulations are needed to derive these new insights. In practice, we found that a few thousand Monte Carlo trials are enough to obtain reasonable approximations. Moreover, the Monte Carlo simulations are embarrassingly parallel and can be efficiently accelerated using multiprocessing with minimal synchronization overhead (Supplementary Figure 17). 

\section*{Discussion}

Currently, there are many open challenges associated with the unsupervised clustering of scRNA-seq data \cite{kiselev2019challenges, stegle2015computational}. First, many clustering methods make strong explicit or implicit prior assumptions about the data. For example, Gaussian mixture models assume that the data is Gaussian distributed. However, previous studies have shown that scRNA-seq data do not follow any known distribution \cite{townes2020quantile}. Second, existing clustering methods cannot internally validate their clustering results. For rare cell type discovery in single-cell analysis, it is crucial to cluster with minimal prior assumptions and report the confidence of the labels at each point for validation. Third, while it is known that discrete cell types exist in single-cell data, some cells can be placed in a continuous gradient of two or more end states without clear boundaries \cite{kiselev2019challenges}. To tackle these existing challenges associated with clustering high-dimensional single-cell data, a clustering algorithm should be specifically designed to meet single-cell analysis needs. 

Here, we developed Forest Fire Clustering inspired by forest fire dynamics \cite{leskovec2007graph}. By reformulating the label propagation algorithm to iteratively find clusters in the data manifold, Forest Fire Clustering offers an efficient and intuitive approach to clustering data. With only one effective hyperparameter that indirectly governs cluster size, our method can discover the number of clusters in the data with minimal prior assumptions. Forest Fire Clustering outperforms previous clustering methods on common benchmarks and shows robust performance on scRNA-seq datasets. Moreover, Forest Fire Clustering can conduct internal validation using Monte Carlo simulations. Clustered labels are randomly propagated to produce point-wise posterior exclusion probabilities, which can quantify the label confidence of each data point and serve as a metric for quality control. Additionally, point-wise label entropies can highlight branching points and key transition cells in developmental pseudo-time. 

Further, Forest Fire Clustering can potentially be used as a soft clustering method. In hard clustering, a data point either belongs to completely to a cluster or not at all, while a data point under soft clustering can fall in multiple categories. Similar to soft clustering labels, the posterior label distribution models the probabilities of a data point being in different clusters. Generating soft clusters that correspond to the hard clusters could help analyze cell types without clear cut-offs in the feature space. Since continuous cell groups can exist in single-cell sequencing data, Forest Fire Clustering is a softer alternative than partitioning continuous cell groups into discrete categories. 

Distribution-based clustering algorithms could be used to produce parameterized label posteriors. However, these label posteriors assume that single-cell data are well distributed. In contrast, our non-parametric label posteriors make minimal assumptions about the shape of the data and are more useful for single-cell analysis. Currently, there are two ways to calculate a non-parametric significance value: bootstrap resampling or permutations. Bootstrap resampling methods (from hierarchical clustering) can only calculate a p-value indicating whether a particular cluster is significant with respect to other data points \cite{Hou2021}. Hence, this significance value is tied to a group of points. In order to compute a confidence measure for each data point, we can rely on permutation-based methods. However, to our knowledge, many other clustering methods are generally limited in generating cluster permutations, or variations of a cluster that are equivalent across rounds of clustering. For example, there is no definitive correspondence between two sets of K-means clustering results because the cluster labels are mathematically invariant. Forest Fire Clustering establishes an equivalence relationship between cluster permutations. New clusters simulated using the seeds from the original cluster are defined as permutations and variations of the original cluster. Then, we can construct a label posterior by attributing the permuted cluster labels to a data point. More importantly, permutation-based methods do not resample from the data and do not suffer from ``double-dipping" issues compared to bootstrap resampling methods \cite{gao2020selective}. 

Many existing methods mimic diffusion or use Monte Carlo for information processing \cite{chen2021diffusion, mixon2017monte}. Indeed, the label propagation in Forest Fire Clustering represents a simplified version of diffusion and is similar to region-growing algorithms for image processing. However, performing repeated label propagations to determine where labels often overlap is unique to Forest Fire Clustering. Crucially, we seek to model self-organized criticality, a concept from statistical physics that governs forest fires, for data mining. Moreover, our approach avoids expensive matrix operations like eigendecomposition on the graph Laplacian, making our methods more scalable and efficient compared to diffusion-based methods. Further, existing graph-based methods rely on objective functions and iterate over the data multiple times. By propagating labels through the data manifold, Forest Fire Clustering makes only one pass through the data, making our approach faster than current graph-based methods. 

As the dimensionality of the data increases, the distances between data points become more uniform, which decreases the signal-to-noise ratio when distinguishing data points from one another. Similar to other clustering methods, Forest Fire Clustering is also limited by the curse of dimensionality. Nevertheless, the impact from the high dimensionality of the data can be reflected in the internal validation of Forest Fire Clustering, as homogenous data points tend to have higher label entropies and posterior exclusion probabilities. Forest Fire Clustering has the advantage of making users aware of the pitfalls in high-dimensional data by providing the validation metrics. In a related way, Forest Fire Clustering is sensitive to high-dimensionality because our approach aims to mimic real fire propagation, which is a low-dimensional process. Hence, Forest Fire Clustering should be performed with non-linear dimensionality reduction (e.g., UMAP, t-SNE, PHATE) to increase separation between data points (Supplementary Figure 15) \cite{mcinnes2018umap, van2008visualizing, moon2019visualizing}.

Guided by clustering and internal validation results, Forest Fire Clustering can be used to discover rare cell types and reveal branching points along developmental trajectories in single-cell experiments. A future goal of single-cell sequencing analysis is to construct a cell-to-cell communication network using interactions between different cell types. Our method offers the possibility of generating robust cell type definitions for network construction and analysis. In addition to single-cell sequencing, our method can be generalized to problems in other domains that involve clustering and evaluating the confidence of the clustered labels. 

\section*{Methods}

\subsection*{Forest Fire Clustering}

Given a set of data points, $x_1, ..., x_n$ each with $m$ features arranged in a data matrix $\pmb W \in \mathbb{R}^{n \times m}$, we calculate the distance matrix $\pmb M \in \mathbb{R}^{n \times n}$ containing the $L^2$ distance between every pair of data points. However, representing the distance matrix for large data sets ($>1$ million data points) is computationally expensive. Therefore, we can approximate the distance matrix by using the $L^2$ distance between the KNNs of each data point. Then, the distances are converted into affinities $\pmb A \in \mathbb{R}^{n \times n}$ using the kernel method. We can use a Gaussian kernel or an adaptive kernel on the KNNs to preserve local distances and encode longer trajectory structures. Similar to preprocessing steps in t-SNE and UMAP, the most suitable kernel bandwidth can be estimated using a binary search on the nearest neighbor distances (e.g., SmoothKNNDist in UMAP). By doing so, a similarity graph $G = (V, E)$ is constructed, with each vertex $v_i$ representing a data point $x_i$ and the corresponding pairwise affinities as edges. The affinity matrix $\pmb A$ acts as the adjacency matrix for graph $G$. In terms of implementing these preprocessing steps, the \textit{scanpy} package in Python has a highly efficient and stable function called ``sc.pp.neighbors" that implements the approximate KNN search and adaptive kernel transformations \cite{wolf2018scanpy}. Hence, we directly use this function in our preprocessing step to obtain the affinity matrix $\pmb A$ from the data matrix $\pmb W$. 

For the label propagation steps, we iteratively cast a label on a vertex and allow the label to propagate until an equilibrium is reached. If one label cannot propagate to all vertices, the procedure is repeated with different labels until all of the vertices in $G$ have been given a label. Each vertex $i$ starts without a label, corresponding to the state $S_i = 0$, and accepts a label $S_i = k$ during the $k$-th label propagation. 

\begin{align}
	S_i = 
	\begin{cases}
		k, & \bar H^k_{labeled, i} \geq T_i\\
		0, & \bar H^k_{labeled, i} < T_i
	\end{cases}
\end{align}

$T_i$ is the threshold above which a vertex would accept a label from neighboring vertices, and the average label influence $\bar H^k_{labeled, i}$ is the average influence from labeled vertices on unlabeled vertex $i$ during the $k$-th label propagation (Equation 4). These values parameterize the label propagation, similar to how they govern forest fire dynamics. There are many hyperparameters in the preprocessing steps such as the K in KNN as well as other parameters involved in the kernel method. It is possible to pair Forest Fire Clustering with other graph construction techniques for preprocessing. Therefore, the core, iterative component of the algorithm has only one effective hyperparameter.

\subsubsection*{Threshold for Accepting Neighboring Labels}

The label acceptance threshold $T_i$ is analogous to flashpoints in forest fire dynamics. A tree in a forest would start burning when the heat surrounding it reaches a certain level, which corresponds to $T_i$ of a vertex $i$ in graph $G$. In order for label propagation to slowly stop at the cluster edges where the data is sparse, $T_i$ should be closely related to the density of the graph. Therefore, $T_i$ is defined to be the inverse of the degree of a vertex $D_i$. As the degree $D_i$ increases, the threshold $T_i$ for accepting a label decreases. Then, label propagation drifts towards regions of high data density when estimating the cluster centers. The threshold is analogous to the activation energy in chemical reactions. 

\begin{align}
	T_i = \frac{1}{D_i}
\end{align}

\subsubsection*{Heat and Fire Temperature}

In a forest fire, heat decays non-linearly with increasing distance and is proportional to the fire temperature. Here, the kernel affinity between two vertices $i$ and $j$ is used to model the non-linear decay in heat as distance increases, and the user-determined hyperparameter $c$ governs the fire temperature. The pairwise label influence between two vertices $i$ and $j$ during the $k$-th label propagation is defined as follows.

\begin{align}
	H^k_{i, j} &= c \cdot A_{i, j}  \cdot \mathbb{I}(k\text{-th label propagation})
\end{align}

Centered on an unlabeled vertex $i$, label influence accumulates from vertices with label $k$ during the $k$-th label propagation. Therefore, the average label influence $\bar H^k_{labeled, i}$ can be modeled as the average pairwise label influence between the unlabeled vertex $i$ and all vertices with label $k$. 

\begin{align}
	\bar H^k_{labeled, i} &=  \frac{ \sum_{j = 1}^{n} \mathbb{I}(S_{j} = k) \cdot H^k_{i, j} }{\sum_{j = 1}^{n} \mathbb{I}(S_{j} = k)}
\end{align}

There are two ways to reach a threshold $T_i$: by setting a higher fire temperature or by using a kernel with a slower decay (or larger bandwidth). When using adaptive kernels, the kernel bandwidth for each vertex is calculated using the distance from its K-nearest neighbors. The affinity matrix $\pmb A$ is better approximated by using higher numbers of nearest neighbors $K$. However, we show experimentally that the parameter K could only marginally increase the clustering quality after a certain threshold (Supplementary Figure 18). An effective hyperparameter is a hyperparameter that influences the size and number of the clusters found. Since the choice $K$ does not play a big role in clustering after a certain threshold, the fire temperature $c$ is the only effective hyperparameter in our clustering algorithm. The fire temperature $c$ plays an important role in clustering and strongly influences the size and number of the clusters discovered, similar to the resolution parameter in Louvain and Leiden. As fire temperature $c$ increases, labels are more likely to propagate to other vertices, causing Forest Fire Clustering to discover larger clusters. On the contrary, smaller clusters are discovered as the fire temperature $c$ decreases. 

For each unlabeled vertex $i$, the clustering algorithm iteratively checks whether the average heat $H^k_i$ is higher than the threshold $T_i$ and adds the vertex to the cluster $k$ if it is. With $k$ clusters and $n$ data points, the algorithm has an approximate worst-case runtime of $O(kn^2)$ when every data point is a cluster. The labels propagate until the following condition is met for every unlabeled point $u$ at the end of the $k$-th label propagation.

\begin{align}
	\{\forall u | S_u = 0, \bar H^k_{labeled, u} < T_u\}
\end{align}

After the $k$-th label propagation reaches a stopping point, another unlabeled vertex will be randomly selected as the starting vertex for the $k+1$-th label propagation. The algorithm continues until all vertices have been labeled, as shown in the pseudo-code of the algorithm below.

\begin{algorithm}[ht]
\SetAlgoLined
\caption{Forest Fire Clustering}
\KwData{Data matrix $\pmb W \in \mathbb{R}^{n \times m}$, fire temperature $c$}
\KwResult{Clustering labels $S_{1...n}$}
    Let $S_{v}$ be a label array for each vertex $v$, initialize values $S_v \leftarrow 0$\;
    Compute pairwise distance matrix $\pmb M \in \mathbb{R}^{n \times n}$ from $W$\;
    Kernel transform $\pmb M$ into affinity matrix $\pmb A \in \mathbb{R}^{n \times n}$\;
    $k \leftarrow 1$ (denoting cluster number)\;
    \While{\normalfont{\# of unlabeled vertices} $>$ 0}{
        Randomly choose a vertex $r$ and set $S_r \leftarrow k$\;
        \For{\normalfont{every unlabeled vertex} $i$}{
            Compute average pairwise heat $\bar H_{labeled, i}^k$\;
            $T_i  \leftarrow  \frac{1}{D_i}$ \;
            \lIf{$\bar H_{labeled, i}^k > T_i$} {$S_i \leftarrow k$} 
        }
    $k \leftarrow k + 1$\;
    }
\end{algorithm}

Since Forest Fire Clustering is a randomized algorithm, we can employ the method of conditional probabilities to improve the stability and lower bound the accuracy when choosing random seeds, similar to K-means++ initialization. Specifically, in the implementation, the first seed is randomly chosen. Instead of randomly choosing the next seed from the remaining points, the next and subsequent seeds are deterministically chosen as the data point that experiences the lowest heat from the previous cluster. Overall, each seed is still chosen randomly, since the first seed is random. However, subsequent seed selections could be seen as pessimistic estimators of the next cluster center. This optimization significantly reduces the stochasticity and increases the stability of Forest Fire Clustering. At the end, our clusters have the following definition. If $k$ is any cluster, then for vertices $\{\forall i | S_{i} = k\}$ and $\{\forall j | S_{j} \neq k\}$:

\begin{align}
	T_{j} &> \frac{ \sum_{i = 1}^{n} \mathbb{I}(S_{i} = k) \cdot H^k_{i,j} }{ \sum_{i = 1}^{n} \mathbb{I}(S_{i} = k) }\\
	&> c \cdot \frac{ \sum_{i = 1}^{n} \mathbb{I}(S_{i} = k) \cdot A^k_{i,j} }{ \sum_{i = 1}^{n} \mathbb{I}(S_{i} = k) }
\end{align}

Each propagation is effectively governed by a local energy function at every vertex that dictates whether the vertex is going to be labeled or not (i.e., ``catch on fire or not"). Once the total heat crosses the threshold $T_i$, it is similar to crossing the activation energy barrier for a chemical reaction, and the heat from the newly labeled vertices (i.e., ``newly burnt tree") contributes to the overall heat and drives continuous label propagation to nearby vertices. The clustering process repeats until the total heat flow cannot cross new activation energy barriers. At that point, the configuration of labeled (i.e., ``on fire") vertices represents a local energy minimum. 

\subsection*{Proving Asymptotic Convergence of Forest Fire Clustering}

In the following section, we show that Forest Fire clusters can asymptotically converge to true cluster distributions up to the second moment. 

Given $n$ $d-$dimensional data points from cluster $k$ $X_1, X_2, ..., X_n \sim N(\mu_k, \sigma_k^2)$ where $\mu_k \in \mathbb{R}^{d}, \sigma_k \in \mathbb{R}^{d \times d}$ such that $||X_1 - \mu_k||^2 < ||X_2 - \mu_k||^2 < ... < ||X_n - \mu_k||^2$, Forest Fire Clustering chooses a random point $X_r$ with heat modeled by the kernel $N(X_r, \sigma^2_{c})$. The kernel bandwidth $\sigma^2_{c}$ is a user-provided hyperparameter for Gaussian kernels or estimated using distances from the KNNs for adaptive kernels. Then for the same arbitrary amount of probability $\epsilon$, the distance $||X_{r+1} - X_r||^2$ is more than $||X_{r-1} - X_r||^2$. 

\begin{align}
	||X_{r+1} - X_r||^2 &> ||X_{r-1} - X_r||^2 \text{ where} \\
	\int^{X_{r+1}}_{X_{r}} \frac{1}{\sigma_k\sqrt{2\pi}}  e^{ -\frac{1}{2}\left(\frac{||x-\mu_k||}{\sigma_k}\right)^{\!2}} dx&= \int^{X_{r}}_{X_{r-1}} \frac{1}{\sigma_k\sqrt{2\pi}} e^{ -\frac{1}{2}\left(\frac{||x-\mu_k||}{\sigma_k}\right)^{\!2}} dx = \epsilon
\end{align} 

Hence, the heat from $X_r$ is higher at $X_{r-1}$ than $X_{r+1}$ under equal probabilities. 

\begin{align}
	N(X_{r+1}; X_r, \sigma^2_{c}) < N(X_{r-1}; X_r, \sigma^2_{c})
\end{align} 

In addition, since $X_{r-1}$ is closer to the true cluster center $\mu_k$, it is located in a denser region under the true probability distribution and is more likely to have a lower label acceptance threshold relative to $X_{r+1}$. Therefore, $X_{r-1}$ is more likely to receive a label from $X_r$ compared with $X_{r+1}$. If $X_{r-1}$ receives the label from $X_r$, the new heat from $\{X_r, X_{r-1}\}$ is modeled by a normal mixture distribution with mean and variance at $\hat \mu_k$ and $\hat \sigma_k^2$.

\begin{align}
\hat \mu &= \frac{X_r + X_{r-1}}{2} \\
\hat \sigma^2 &= \frac{\sigma^2_c + \hat \mu_k^2 - \mu_k^2}{2} + \frac{\sigma^2_c + \hat \mu_k^2 - \mu_k^2}{2} = \sigma^2_c + \hat \mu_k^2 - \mu_k^2
\end{align}

If the fire temperature $c$ is selected such that $\sigma_{c}^2 \simeq \sigma_k^2$ and heat can cross the threshold of all points $X_1 ... X_n \sim N(\mu_k, \sigma_k^2)$, then the mean and variance approaches the expected value and variance of the true cluster distribution by the Law of Large Numbers. Intuitively, Forest Fire Clustering can be understood as a specific instance of kernel density estimation (KDE) applied to clustering. Even though each Gaussian kernel assumes a certain shape, KDE is still an effective way to non-parametrically estimate the underlying distribution. Similarly, FFC can also non-parametrically approximate the global cluster distribution in the data by applying the kernel to each data point.

\begin{align}
\hat \mu = \frac{X_1 + ... +  X_n}{n} \rightarrow \mu_k \\
\hat \sigma^2 = \frac{\sigma^2_c + ... +  \sigma^2_c}{n} \rightarrow \sigma_k^2
\end{align}

Forest Fire Clustering iteratively labels clusters $1...k$, and data points are added to each cluster in a greedy fashion. Hence, Forest Fire Clustering can approximate the underlying data distributions conditioned on the user-provided fire temperature $c$. Further, Forest Fire Clustering can be seen as kernel density estimation applied to clustering. With kernel affinities and convergence by the Law of Large Numbers, the argument can also be generalized to cases where the true cluster distribution is not Gaussian.

\subsection*{Monte Carlo Validation}

After performing Forest Fire Clustering, the initial clustering labels $S_{1...n}$ for all $n$ vertices are stored. Internal validation can be performed on each initial clustering label by constructing a posterior label distribution. In the $t$-th Monte Carlo trial, a random seed vertex $r$ is selected, and the seed vertex label $S_{r, t}$ at trial $t$ takes on the same label $S_{r}$ as it did in the initial clustering results. Labels on all other vertices are cleared. The seed label $S_{r, t}$ is allowed to propagate until the average label influence can no longer cross the threshold of the remaining unlabeled vertices. For all vertex $i$ that accepted the seed label $S_{r, t}$ in trial $t$, it adds one occurrence of that seed label in their records $S_{i, t} = S_{r, t}$. After $T$ Monte Carlo trials, at each data point $i$, the accepted seed labels $S_{i, 1...T}$ are tabulated to construct a discrete label frequency distribution, which we call the posterior label distribution. The probability that a data point $i$ being labeled as cluster $k$ could be calculated by $P^k_i = \frac{1}{T} \sum^T_{t = 1} \mathbb{I}(k = S_{i, t})$. With the simulated posterior probability distribution, we can compute point-wise posterior exclusion probabilities with respect to the original label and point-wise entropies across all labels. The posterior exclusion probability of a data point $i$ can be calculated by:

\begin{equation*}
	PEP_i = 1 - \sum^{\{k \neq S_i\}}_{k} P^k_i
\end{equation*}

The entropy of a data point $i$ can be calculated by:

\begin{equation*}
	E_i = -\sum^k P^k_i \log(P^k_i)
\end{equation*}

Merging non-intersecting clusters from different Monte Carlo trials into a complete set of cluster labels forms a permutation of the labels on the data. The number of Monte Carlo trials varies with the number of clusters in the data, as a larger number of clusters need more Monte Carlo trials to obtain proper approximation resolution. A large number of Monte Carlo trials can be computationally intensive, but the computation can be efficiently parallelized with shared memory multi-threading. We found that 2,000 to 10,000 trials obtained reasonable approximations for datasets presented in this paper. 

There is an edge case where Monte Carlo internal validation could not accurately estimate the uncertainty. When there is only one cluster in the data, every data point can only receive one type of label. In single-cell sequencing, it is not realistic to discover only one cell type in the data, therefore the practicalities of single-cell sequencing could guide our analysis away from this edge case.

\begin{algorithm}[ht]
\SetAlgoLined
\caption{Monte Carlo Validation}
\KwData{Affinity matrix $\pmb A \in \mathbb{R}^{n \times n}$, fire temperature $c$, initial label $S_{1...n}$, total trial number $T$}
\KwResult{Posterior exclusion probability $PEP_{1...n}$, entropy $E_{1...n}$}
    Let $S_{v, t}$ be a label matrix for vertex $v$ at trial $t$, initialize values $S_v \leftarrow 0$\;
    $k \leftarrow 1$ (denoting cluster number)\;
    \For{$t \leftarrow 1...T$}{
        Randomly choose a vertex $r$ with initial label $S_r$, set $S_{r, t} \leftarrow S_r$\;
        \While{\normalfont{\# of unlabeled vertices} $>$ 0}{
            Calculate average label influence $\bar H^k_{labeled, i}$\;
            $T_i  \leftarrow  \frac{1}{D_i}$\;
            \lIf{$\bar H_{labeled, i}^k > T_i$} {$S_{i, t} \leftarrow S_{r, t}$} 
        }
    }
 $P^k_i \leftarrow \frac{1}{T} \sum^T_{t = 1} \mathbb{I}(k = S_{i, t})$\;
 $PEP_i \leftarrow 1 - \sum^{\{k \neq S_i\}}_{k} P^k_i$\;
 $E_i \leftarrow -\sum^k P^k_i \log(P^k_i)$\;
\end{algorithm}

Other Python libraries were used to reduce runtime. For example, the numpy package was used for linear algebra operations \cite{harris2020array}. The scipy package was used for sparse matrix acceleration \cite{2020SciPy-NMeth}, and the numba package was used for numerical acceleration \cite{lam2015numba}.

\subsection*{Analysis of Simulated Data}

Gaussian mixtures were generated by sampling from Gaussian distributions evenly centered on the unit circle. Two Gaussian bandwidths were used in the analysis, $\sigma=0.15$ and $\sigma=0.2$. Both sets of Gaussian mixture data were clustered with $K=50$ and fire temperature $c=50$. For online clustering, a dataset with four clusters was simulated and clustered with Gaussian kernel $\sigma=0.2$ and fire temperature $c=10$. Then, another dataset with eight clusters (four of which were from the same distribution as the previous dataset) was simulated and clustered based on previous labels using the same set of parameters. The performance was evaluated on the out-of-sample testing data points. Synthetic data of different shapes were produced by the sklearn package in Python \cite{scikit-learn}. The other clustering algorithms in the benchmark were selected because they are representative and widely used clustering algorithms implemented in the sklearn package. In general, clustering parameters for alternative methods were chosen to generate around the same number of clusters as the ground truth. For example, discrete cell groups were clustered with $K=5$ for SC3 and resolution$=0.5$ for Louvain. 

Simulated scRNA-seq data were generated using the R package Splatter. In total, 1k cells and 5k genes were simulated for 20 datasets. After normalization and standardization, the first 30 principal components were used for clustering. For discrete cell group simulations, five groups were generated $(0.45, 0.15, 0.15, 0.15, 0.1)$ with parameters $(out.prob = 0.20, out.facScale=1.5)$, and the cells were clustered with fire temperature $c=50$ and $K=50$. For continuous cell group simulations, four groups were generated $(0.35, 0.25, 0.25, 0.15)$ with parameters ($out.prob = 0.1$, $out.facScale=0.25$, $de.prob = 0.5$, $de.facLoc = 0.2$) along the differentiation path $(0,0,0,0)$, and the cells were clustered with $K=50$ and fire temperature $c=30$. For visualizations, Savitzky-Golay filters were used to fit local polynomials and highlight the inverse relationship between developmental pseudo-time and label entropy. 

There are three main ways to evaluate the performance of unsupervised methods. First, purity scores evaluate the cluster labels based on a set of references labels. To calculate the purity score, one can construct a confusion matrix and assign the dominant predicted cluster to each ground truth cluster. Then, the sum of the total data points in dominant predicted clusters is divided by the total number of data points to compute the purity score. The metric quantifies whether the predicted clusters contain only points from a reference cluster by a value from zero to one. Second, the ARI also measures the similarity between two clusterings. The Rand index quantifies whether all pairs of points in two clusterings match up to the same cluster. The ARI is the Rand index adjusted for chance. Silhouette scores evaluate the similarity of data points within a cluster compared with the similarity of data points outside of the cluster. Different from the purity score and ARI, the silhouette score does not require ground truths as a reference to compute the statistic. 

\subsection*{Analysis of Experimental Data}

The RNA gene expression count data for both PBMC datasets were preprocessed using the ``zheng17" recipe in the scanpy package in Python \cite{zheng2017massively}. For CITE-seq, cell surface proteins were used to label PBMC cell types using marker genes available from \href{https://pages.10xgenomics.com/rs/446-PBO-704/images/10x_PS032_SCGE_Protein_Single_Cell_with_Surface_Protein_digital.pdf}{10x Genomics}. The first 30 principal components were used to generate other low-dimensional embeddings (UMAP, t-SNE, Diffusion Maps, PHATE). For graph-based methods (like Louvain, Leiden, and PARC), $K=50$ was used to generate a KNN graph from the low-dimensional embeddings. Additionally, the K-sparse affinity matrix used by Forest Fire Clustering was constructed from the same KNN graph. Hence, clustering was performed on the low-dimensional embeddings with different hyperparameter configurations (with resolution$=(0.05, 0.1, 0.5, 1, 5)$ for Louvain, Leiden, and PARC, $K=(2, 4, 6, 8, 10)$ for K-means, and $c=(20, 40, 60, 80, 100)$ for Forest Fire Clustering). To generate a stronger baseline for benchmarking, the best performance across the hyperparameter configurations and dimensionality reduction methods was recorded for each clustering method.

The clustering results were evaluated using cell surface protein labels. To generate a stronger baseline for benchmarking, the best performance across the hyperparameter configurations and dimensionality reduction methods was used for comparison between clustering methods. Leiden clusters with \textit{Procr} marker expression were used as roots for diffusion pseudo-time analysis. For multiomic sequencing, cell type labels were inferred from RNA gene expression count using the pegasus package in Python with the ``pg.infer\textunderscore cell\textunderscore types" function and the ``human\textunderscore immune" marker gene set \cite{Li823682}. The ATAC peak counts were log-transformed and standardized, and the top 10,000 highly variable ATAC peaks were selected for PCA. Similarly, the first 30 principal components were used to generate other low-dimensional embeddings (UMAP, t-SNE, Diffusion Maps, PHATE). For graph-based methods (like Louvain, Leiden, and PARC), $K=50$ was used to generate a KNN graph from the low-dimensional embeddings. Additionally, the K-sparse affinity matrix used by Forest Fire Clustering is constructed from the same KNN graph. In PBMC, our analysis shows that UMAP embeddings obtained the best quality Forest Fire clusters compared to other dimensionality reduction methods (Supplementary Figure 19). Previous works have also supported UMAP embeddings as effective coordinates for clustering \cite{mcinnes2018umap}. 

The gene expression count matrix for MCA was also preprocessed using the ``zheng17" recipe in the scanpy package but with the top 5000 highly variable genes selected. Similarly, the first 30 principal components were used to generate other low-dimensional embeddings (UMAP, t-SNE, Diffusion Maps, PHATE). For graph-based methods (like Louvain, Leiden, and PARC), $K=50$ was used to generate a KNN graph from the low-dimensional embeddings. Additionally, the K-sparse affinity matrix used in Forest Fire Clustering is constructed from the same KNN graph. Therefore, clustering was performed on the these low-dimensional embeddings with different hyperparameter configurations (with resolution$=(0.05, 0.1, 0.5, 1, 5)$ for Louvain, Leiden, and PARC, $K=(2, 4, 6, 8, 10)$ for for K-means, and $c=(20, 40, 60, 80, 100)$ for Forest Fire Clustering). In MCA, t-SNE embeddings (with $perplexity=75$) produced better separation between cell types than other methods. Therefore, t-SNE embeddings were used for visualizations. The tissue labels were used as ground truths for benchmarking. Differentially expressed genes were extracted by using scanpy’s ``rank\textunderscore genes\textunderscore groups function" with the Wilcoxon test and Bonferroni corrections. Error bars were generated by repeating each experiment ten times. 

\subsection*{Online Clustering}

In certain scenarios, it is more convenient to make inferences on a small number of new data points without re-clustering, especially when the number of data points is initially large ($>1$ million cells), also known as the ``n+1" clustering problem. Therefore, we formulated Forest Fire Clustering to allow for the continuous integration of a few new out-of-sample data points. The dataset previously clustered can be seen as the training data, and the newly arrived data can be viewed as the testing data. The goal of online clustering is to correctly classify the cluster labels in the testing data, which may contain new clusters. Data preprocessing and clustering are first performed on the training data. Then, we independently normalize and log-transform the testing data, as if they arrived at different times. Further, we assume that the training and testing data will share the same set of features. If the features of the training and testing data are disjoint, then it would not be possible to extrapolate clusters from the training data to the testing data. Therefore, we subset the testing data to contain only the top genes from the training data as features, and we correct batch effects between the training and testing data with mutual nearest neighbor batch correction. Then, the testing data are projected onto the low-dimensional representations of the training dataset. A joint affinity matrix of the training and testing data is then calculated from the joint low-dimensional embeddings. After the joint affinity matrix is computed, we then extend the cluster predictions to the testing data.

For each data point in the testing data, Forest Fire Clustering checks whether the average label influence from the clustered data can cross the threshold of the testing data point. If there are multiple label influences that can cross the threshold, then the testing data point takes on the label with the highest influence. If there are no label influences that can cross the threshold, then testing data point becomes a seed vertex for a new cluster. Therefore, after seeds for new clusters are selected in the testing data, Forest Fire Clustering propagates the new labels through the testing data manifold, allowing new clusters to be discovered in the testing data. 

\section*{Data Availability}

Synthetic data were simulated using sklearn (in Python), and scRNA-seq data were simulated with Splatter (in R). Experimental scRNA-seq data were downloaded from existing public repositories. CITE-seq and multiomic sequencing of PBMC datasets are available from \href{https://www.10xgenomics.com/resources/datasets}{10x Genomics}. The Mouse Cell Atlas dataset is also publicly available to download from \href{http://bis.zju.edu.cn/MCA/}{their website}. 

\section*{Code Availability}
Tutorials and all source code for Forest Fire Clustering are available on Github at:
\url{https://github.com/gersteinlab/forest-fire-clustering} or \url{https://zenodo.org/badge/latestdoi/432892882}.


\section*{Acknowledgments}
Research reported in this publication was supported by the National Institutes of Health under award numbers K01MH123896, U01DA053628, and UM1DA051410.

\section*{Author Contribution Statement}
Zhanlin Chen and Philip Tuckman formulated the algorithm. Zhanlin Chen and Jeremy Goldwasser implemented the algorithm. Zhanlin Chen and Jason Liu gathered data and performed benchmarking. Jing Zhang and Mark Gerstein served as co-advisors and supervised the research. Zhanlin Chen wrote the manuscript with input from all authors. All authors read and approved the final manuscript.

\section*{Competing Interests Statement}
The authors have no conflicting interests to declare.


%
%
%

\end{document}